\documentclass{article} 
\usepackage[dvipsnames]{xcolor}
\usepackage{iclr2021_conference,times}

\usepackage{amsmath,amsfonts,bm}

\def\eqref#1{equation~\ref{#1}}

\def\1{\bm{1}}

\DeclareMathAlphabet{\mathsfit}{\encodingdefault}{\sfdefault}{m}{sl}
\SetMathAlphabet{\mathsfit}{bold}{\encodingdefault}{\sfdefault}{bx}{n}

\newcommand{\R}{\mathbb{R}}

\usepackage{hyperref}
\usepackage{url}
\usepackage{tikzdefs}

\usepackage{makecell}
\usepackage{multirow}
\usepackage{pgfplotstable}
\usepackage{booktabs}
\usepackage{xparse}
\usepackage{float}
\usepackage{cleveref}
\usepackage[font=footnotesize,justification=centering]{caption}
\newcommand{\XXX}[1]{\textcolor{green}{\\XXX: #1\\}}
\renewcommand{\vec}{\mathbf}

\newcommand{\set}{\mathcal}

\newcommand{\loss}{\textit{L}}
\newcommand{\lossCC}{\textit{L}^\textit{CC}}
\DeclareMathOperator{\mean}{\mu}
\DeclareMathOperator{\var}{\sigma^2}

\NewDocumentCommand{\photostrip}{ O{0} O{0} m m }{
    \centering
    \normalsize{#3}
    \\[-5pt]
    \rule{0.4\textwidth}{.4pt}
    \par
    \centerline{
    \fcolorbox{white}{white}{
    \includegraphics[width=\textwidth,trim={#1 0 #2 0},clip]{#4}}}
    \vspace*{10pt}
}

\usepackage{subfig}

\title{Explicit Domain Adaptation through implicit knowledge: Post-hoc Data Homogenization}

\title{Guided Data Homogenization for Post-Hoc Domain Adaptation}

\title{Post-Hoc Domain Adaptation via Guided Data Homogenization}

\author{%
Kurt Willis \& Luis Oala\\
Fraunhofer HHI\\
Berlin, Germany\\
\texttt{\{kurt.willis,luis.oala\}@hhi.fraunhofer.de} \\
}

\iclrfinalcopy 
\begin{document}

\maketitle

\begin{abstract}
Addressing shifts in data distributions is an important prerequisite for the deployment of deep learning models to real-world settings. A general approach to this problem involves the adjustment of models to a new domain through transfer learning. However, in many cases, this is not applicable in a post-hoc manner to deployed models and further parameter adjustments jeopardize safety certifications that were established beforehand. In such a context, we propose to deal with changes in the data distribution via guided data homogenization which shifts the burden of adaptation from the model to the data. This approach makes use of information about the training data contained implicitly in the deep learning model to learn a domain transfer function. This allows for a targeted deployment of models to unknown scenarios without changing the model itself. We demonstrate the potential of data homogenization through experiments on the CIFAR-10 and MNIST data sets. 
\end{abstract}



A routine assumption in machine learning is that data 
in the training and testing environment is identically distributed. 
In reality, data distributions may differ \citep{hendrycks2020faces}, leading to performance degradations for models trained with 
standard deep learning algorithms.
Transfer learning \citep{XFER_SURVEY} has emerged as a field
that aims to transfer knowledge of models from a learned source task 
to a target task by exploiting underlying commonalities.
Domain adaptation in particular,
addresses a mismatch in either
the input space $\mathcal X$ 
or the data distribution $\mathbb P$
of a source and target task:
$(\mathcal X_s, \mathbb P_s) \neq (\mathcal X_t, \mathbb P_t)$.
A common approach to this challenge is to encourage the model to learn a domain invariant
representation of the data.
This is typically done through the use of the joint distribution, 
requiring access to the source data set.

This work
proposes domain adaptation through 
homogenization of the data in a post-hoc-manner.
This eliminates the need to anticipate numerous possible test-time scenarios during training
and
the original model is kept intact - a desirable property for security and robustness certification \citep{neurips2019_f7fa6aca,pmlr-v136-oala20a}. 
In detail, an explicit domain mapping function is learned 
through an optimization objective
adopted from work by \cite{DeepInversion}.
This does not assume access to the source data set;
instead, the data statistics measured in an appropriate feature space are sufficient.
These statistics are readily available in networks that make use of
the widely adapted batch normalization layers \citep{ioffe2015batch}.

An example for the relevance of post-hoc domain adaptation is seen in optical systems which provide inputs for deep learning applications. Sensor corrosion or re-calibration of hardware devices has proven to be a significant barrier to the reliable application of deep learning systems in fields such as medicine \citep{google} or autonomous driving \citep{michaelis2020benchmarking}.

We show the ability to homogenize non-identically distributed data 
in experiments on the CIFAR-10 \citep{CIFAR} and MNIST \citep{MNIST} data sets.
The method is further tested in the unsupervised setting,
where no labeled data in the target domain is available.

Related work focuses on learning domain-invariant representations
in order to bridge the discrepancy in distributions.
In this effort, \cite{da_ae} propose training an auto-encoder that is able to encode both domains.
\cite{da_deep_transfer} make use of the criterion loss along with a domain confusion loss 
which is implemented by adding an adaptation layer on top of the classifier's last layer.
This auxiliary layer aims to make the learned representations indistinguishable 
between domains by maximally confusing a domain classifier.
\cite{da_mmd} further extend the use of adaptation layers to multiple layers throughout the network.
\cite{da_cmd} and \cite{deep_coral} 
formulate a distribution loss incorporating higher-order moments.
\cite{da_prototypes} propose to classify data of the target domain by comparing 
feature representations to prototypes for each class.
\cite{da_cyclegan} propose CycleGAN, a generative adversarial network that learns 
explicit domain transformations, by formulating an adversarial loss
and measuring reconstruction accuracy.

This work can also be viewed in the context of inverse problems,
and more specifically, related to the task of deblurring or deconvolution.
Whereas \cite{IP_Deconv} and \cite{IP_VN} 
train a deconvolution model for
image reconstruction on handcrafted perturbation functions, 
this work solely makes use of the prior knowledge
learned by an image classification network and statistics
to correct for general distribution shifts
that also entail deconvolutions. 


\section{Method}
\label{sec:method}

\begin{figure}
    \centering
    \scalebox{0.9}{
        \input{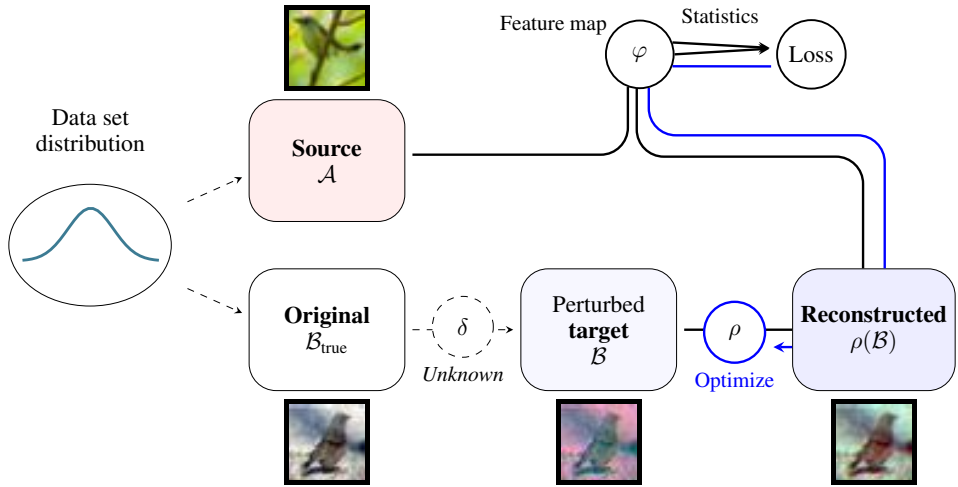}
    }
    \caption{Overview of the homogenization setting. A general transformation $\rho$ is optimized, so that a neural network $\Phi$ regains its performance on the transformed dataset $\rho(\set B)$. The feature maps are obtained from $\Phi$.}
    \label{fig:outline}
    \centering
\end{figure}
\Cref{fig:outline} summarizes the main idea of guided data homogenization.
Suppose we are given 
a neural network that was trained on - or performs reasonably well on - a source data set $\set A$
and 
a target data set $\set B$, typically smaller in size ($n_{\set B} < n_{\set A}$).
Further,
the distribution of $\set B$ differs crucially from $\set A$, 
i.e. the neural network is unable to attain high performance 
on data coming from $\set B$.
The task is to learn a transformation $\rho$ that adjusts $\set B$ 
so that the network's performance is regained.

An underlying assumption is that $\set B$ originates from the same distribution
as $\set A$, but has been corrupted or transformed by an unknown perturbation $\delta$.
The transformation $\rho$ is modeled to correct for this perturbation, 
in the sense that $\rho(\set B) \approx \set B_\text{true}$.
This is done by minimizing a dissimilarity-score between $\rho(\set B)$ and $\set A$ 
in the feature space given by an appropriate feature map $\varphi$.
Towards this goal, simple statistics (mean and variance)
are recorded for both data sets after they have been transformed by $\varphi$.
By use of backpropagation, $\rho$ is optimized, 
in order to make the statistics 
of the target $\set B$ approach those of the source $\set A$.
Whether matching statistics in feature space actually increases similarity between $\set A$ and $\set B$
strongly depends on $\varphi$ and its representation of the data set.
%

For a given feature mapping $\varphi:\R^d \to \R^m$, 
the dissimilarity score, or the \textbf{statistics-loss}, is defined as follows.
\begin{equation}
    \loss _{\varphi} (\set A, \set B) = 
    \|\mean ({\varphi (\set A)}) - \mean ({\varphi (\set B)})\|_2 +
    \|\var ({\varphi (\set A)}) - \var ({\varphi (\set B)})\|_2
    \label{eqn:statistics_loss}
\end{equation}
$\mean$ and $\var$ denote the empirical feature mean and variance over the feature channels,
as is generally done in the computations of BatchNorm-layers.
%
A data set mapping is obtained by applying a feature map $\varphi$ to every input:
$
    \varphi(\set A) := \{(\varphi(\vec x), y) \mid (\vec x, y) \in \set A\} \;
$.
This notion can be extended to a collection of feature maps:
$
    \loss_{[\varphi_1, \dots, \varphi_n]} (\set A, \set B) = 
    \sum_{i=1}^n \loss_{\varphi_i} (\set A, \set B)
$.
Furthermore, a \textbf{class-dependent} loss can be obtained by splitting the data set into 
subsets for each class and accumulating the resulting losses:
$
    \lossCC _\varphi (\set A, \set B) =
    \sum _{c = 1}^C
    \loss_\varphi(\set A|_c, \set B|_c) \, .
$
Here, $\set A|_c = \{(\vec x, y) \in \set A \mid y = c\}$ is
the subset of $\set A$ constrained to samples of label $c$.
This case can be compared to related work, where class-prototypes are used.

\section{Experiments}
\begin{figure}
    \centering
    
    \subfloat[][Data set $\set A$]{\includegraphics[trim={0 0 22.19cm 0},clip,width=0.30\textwidth]{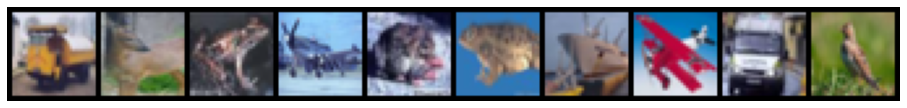}}
    \subfloat[][Data set $\set B$]{\includegraphics[trim={0 0 22.19cm 0},clip,width=0.30\textwidth]{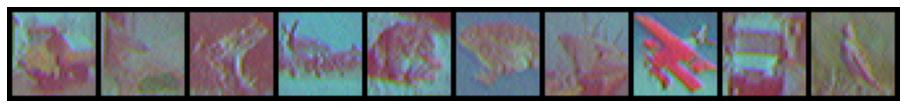}}
    \subfloat[][Homogenized $\set B$]{\includegraphics[trim={0 0 22.19cm 0},clip,width=0.30\textwidth]{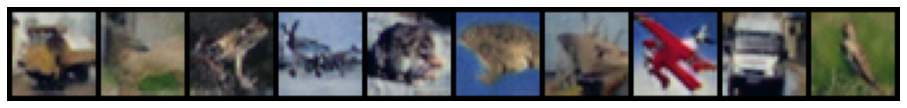}}
    \caption{Excerpt of results on the homogenization experiments on CIFAR10 data set in a supervised setting 
    after 100 epochs for NN ALL method. $\kappa=0.15$, $n_{\set B}=512$. The full results are depicted in  \Cref{app:CIFAR_Images} of the Appendix.}
    \label{fig:CIFAR_Images}
    \vspace{-2em}
\end{figure}

The experiments\footnote{The code for guided data homogenization and all experiments can be accessed at \url{https://github.com/willisk/Thesis}.} are conducted on the popular image recognition data sets
MNIST and CIFAR-10.
The target data set $\set B$ is obtained by applying
a randomized, parameter-controlled perturbation 
to a small unseen subset of the original data set.
The \textbf{perturbation model} 
is a composition of additive Gaussian noise and two noise-controlled 2d-convolutions:
$
    \delta(\vec x) = \vec K_\kappa^{(2)}(\vec K_\kappa^{(1)}(\vec x + \kappa \boldsymbol \mu)
$.
The kernel matrices of the convolutions are each of size $3\times3$ and initialized to the identity kernel with added Gaussian noise.
The overall noise level is controlled by the parameter $\kappa$.
The \textbf{reconstruction model} 
is a vanilla four-layer convolutional residual network of width 16.


%

\subsection*{Feature maps}
As explained in \Cref{sec:method}, the statistics-loss depends on the feature mapping $\varphi$. Various feature mappings are compared in the experiments.
The main approach (\textbf{NN ALL}) involves
all hidden states of a neural network $\Phi = (\Phi_\text{L} \circ \dots \Phi_1)$.
These are seen as the outputs of a collection of feature maps [$\varphi_0, \dots, \varphi_\text{L-1}$], where
\begin{alignat*}{2}
    \varphi_\ell &: \R^d \to \R^{d_\ell} &&= (\Phi_\ell \circ \dots \Phi_1)
    \,, \textnormal{ for } \ell = 1, \dots, L - 1   \\
    \varphi_\textnormal{0} &: \R^d \to \R^d &&= \text{id} \,.
\end{alignat*}

Another feature mapping serving as a comparison is
induced by 
the penultimate layer $\Phi_{\text{L}-1}$, 
the layer before the logits layer.
It is reported as \textbf{NN}.
\[
    \varphi : \R^d \to \R^{d_\textnormal{L-1}} = (\Phi_\textnormal{L-1} \circ \dots \circ \Phi_1)
\]

In order to further delineate the benefit
of using representations stemming from
the neural network,
a feature mapping involving only one linear layer with a width of 512 is considered.
Comparing the performance of the reconstruction using shallow 
versus deep features can further warrant
the necessity of complex feature maps.
This one linear layer can be seen as a set of random linear projections; it is reported as \textbf{RP}.

The hidden representations of the same neural network model with randomly initialized parameters (\textbf{RANDOM NN}) serves as another baseline comparison.
This is done in order to test the utility of an optimized feature representation.

The complete loss formulation 
for the supervised setting is then given by
\[
    r_\text{stats}\loss_\varphi(\set A, \set B) + r_\text{crit}\loss_\text{crit}(\set B, y_{\set B}) \,,
\]
where $y_{\set B}$ are the target labels and $\loss_\text{crit}$ is the criterion-loss, 
which, in this case, is the cross-entropy loss.
The unsupervised setting only makes use of $\loss_\varphi$ (i.e. $r_\text{crit}=0$).



\section{Results}

For measuring the overall distortion of the reconstruction,
the \textbf{l2-error} $\varepsilon_2$ of the unit vectors is calculated. 
$
\varepsilon_2^2 = \frac 1 d \sum_{i=0}^d \|\rho (\delta (\vec e_i)) 
    - \vec e_i  \|_2^2 \,,
$
where $\vec e_i$ is the i-th unit vector of the standard basis.
Further, 
a metric commonly encountered in image quality assessment, the 
\textbf{peak signal-to-noise ratio} (\textbf{PSNR}) is reported.
One last metric, that doesn't rely on the mean-squared error,
the \textbf{structural similarity index measure} (\textbf{SSIM}) \citep{PSNRvsSSIM} is considered. 
It is re-scaled to fit the interval $[0, 1]$ and is given as a percentage (\%).

In order to measure the generalization ability 
of the learned transformation,
a second, independently trained neural network's ($\Phi_\text{ver}$) accuracy is reported
as the \textbf{verification accuracy}. 
Further, the same transformations are applied to an unseen validation set $\set C$.
The accuracy attained by $\Phi_\text{ver}$ on this new set 
constitutes the \textbf{validation accuracy}.

The baseline evaluation metrics, 
along with the final results of the reconstruction task 
on the CIFAR-10 data set 
are given in \Cref{tab:cifar10results} for each method.
Less than $1\%$ of the original training set sample size is used
in these experiments.
After the perturbation has been applied to $\set B$
the scores drop significantly.
Relying solely on the criterion 
results in overfitting the reconstruction function,
as is seen by the disparity in generalization scores.
Random projections (RP) and the randomly initialized convolutional neural network
fail to capture vital information about the data set distribution;
although they are still able to improve various scores.
The best-performing method is NN ALL.

While shallow features are able to perform the reconstruction task on simple data sets
like MNIST (see \Cref{tab:MNIST_results} of the Appendix),
their benefit on a more complex data set like CIFAR-10 is limited.
Further results for the class-conditional formulation $\lossCC$ 
are given
in \Cref{app:fullresults}.
Using this formulation, a significant performance boost is noted
when $r_\text{crit}=0$, however, 
no notable benefit is obtained
when the criterion loss is included.

The evaluation metrics for the unsupervised setting, 
where no label information is provided
are given in \Cref{tab:cifar10results_unsupervised} of the Appendix.
The scores are close to the supervised case and
show promising results even for small sample sizes.


\begin{table}
\vspace{-2em}
\centering
\caption{Metrics of homogenization results in the supervised setting after 100 optimization epochs for CIFAR-10. The mean and the standard deviation over 5 runs are given. The number of samples $n_{\set B}=512$.}
\scalebox{0.93}{
\footnotesize
\begin{filecontents*}{figures/multiple_CIFAR10_baseline_results.csv}
baseline,acc,acc(val),acc(ver),l2-err,PSNR,SSIM
Source A,99.7,,99.8,0,inf,100
Target B (original),99.8,93.2,99.8,0,inf,100
Target B (distorted),26.2,26.6,26.0,24.3,16.5,84.0
CRITERION,99.6 +/- 0.4,77.9 +/- 2.1,75.9 +/- 2.7,14.0 +/- 2.7,18.0 +/- 0.4,86.7 +/- 2.2
NN,95.5 +/- 0.6,83.1 +/- 0.6,80.8 +/- 0.8,11.6 +/- 1.6,19.5 +/- 0.3,90.1 +/- 0.5
NN CC,99.9 +/- 0.1,85.8 +/- 0.7,83.0 +/- 0.6,9.9 +/- 2.2,20.9 +/- 0.6,92.0 +/- 0.9
NN ALL,99.3 +/- 0.4,88.2 +/- 0.6,84.9 +/- 1.0,10.0 +/- 2.6,23.3 +/- 0.3,95.4 +/- 0.2
NN ALL CC,99.8 +/- 0.0,87.8 +/- 0.9,84.7 +/- 1.0,12.8 +/- 0.9,23.7 +/- 0.5,95.4 +/- 0.4
RANDOM NN,51.9 +/- 5.0,48.4 +/- 5.7,42.7 +/- 4.2,29.0 +/- 10.6,11.3 +/- 0.7,59.6 +/- 4.7
RANDOM NN CC,55.6 +/- 1.6,53.0 +/- 2.7,48.1 +/- 1.9,38.2 +/- 10.8,13.5 +/- 0.6,67.4 +/- 4.3
RP,51.1 +/- 2.1,51.0 +/- 1.4,47.3 +/- 1.1,76.4 +/- 5.2,18.1 +/- 0.2,82.6 +/- 0.6
RP CC,42.6 +/- 0.9,42.5 +/- 0.9,41.4 +/- 1.2,42.2 +/- 1.7,17.0 +/- 0.2,81.5 +/- 0.4
\end{filecontents*}
\pgfplotstabletypeset[
baseline,
display columns/0/.style={column name=\textbf{Data set}, column type=l, string type},
images,
skip rows between index={5}{6},
skip rows between index={7}{8},
skip rows between index={9}{10},
skip rows between index={11}{12},
every row no 3/.style={
    before row={\\[-5pt] \textbf{Homogenized $\set B$} \\\midrule}
},
every row 3 column 1/.style={highlight},
every row 6 column 2/.style={highlight},
every row 6 column 3/.style={highlight},
every row 6 column 4/.style={highlight},
every row 6 column 5/.style={highlight},
every row 6 column 6/.style={highlight},
]{figures/multiple_CIFAR10_baseline_results.csv}
}
\label{tab:cifar10results}
\end{table}

    
    




\section{Conclusions}

Neural networks, and convolutional neural networks in particular for image data sets,
are powerful feature extractors.
The learned feature representations carry implicit knowledge about the training data,
and by tracking simple statistics of the latent states, 
more information about the data distribution is gained.
\cite{DeepInversion} have shown that it is possible to recover high-fidelity,
plausible images from only having access to the BatchNorm-statistics.
This work demonstrates that it is also possible to use this information
in the context of domain adaptation for data homogenization
and, in particular, 
shows its applicability to deconvolution or re-calibration tasks.
Guided data homogenization proves to be viable even 
when few data in the target domain is available 
or
even when no label information in the target domain is given.
Further exploration of the connection between implicit model knowledge and post-hoc model adaptation offers an intriguing avenue for future research.

\bibliography{references}
\bibliographystyle{iclr2021_conference}

\newpage

\appendix
\section{Full results}
\label{app:fullresults}
\subsection{CIFAR-10}
\label{app:cifar}

\begin{table}[h]
\centering
\caption{Metrics on homogenization results in the supervised setting after 100 optimization epochs for CIFAR-10, including the class-conditional formulation (CC). The mean and the standard deviation over 5 runs are given. $n_{\set B}=512$.}
\scalebox{0.93}{
\footnotesize
\begin{filecontents*}{figures/multiple_CIFAR10.csv}
Methods,acc,acc(val),acc(ver),l2-err,PSNR,SSIM
CRITERION,99.6 +/- 0.4,77.9 +/- 2.1,75.9 +/- 2.7,14.0 +/- 2.7,18.0 +/- 0.4,86.7 +/- 2.2
NN,95.5 +/- 0.6,83.1 +/- 0.6,80.8 +/- 0.8,11.6 +/- 1.6,19.5 +/- 0.3,90.1 +/- 0.5
NN CC,99.9 +/- 0.1,85.8 +/- 0.7,83.0 +/- 0.6,9.9 +/- 2.2,20.9 +/- 0.6,92.0 +/- 0.9
NN ALL,99.3 +/- 0.4,88.2 +/- 0.6,84.9 +/- 1.0,10.0 +/- 2.6,23.3 +/- 0.3,95.4 +/- 0.2
NN ALL CC,99.8 +/- 0.0,87.8 +/- 0.9,84.7 +/- 1.0,12.8 +/- 0.9,23.7 +/- 0.5,95.4 +/- 0.4
RANDOM NN,51.9 +/- 5.0,48.4 +/- 5.7,42.7 +/- 4.2,29.0 +/- 10.6,11.3 +/- 0.7,59.6 +/- 4.7
RANDOM NN CC,55.6 +/- 1.6,53.0 +/- 2.7,48.1 +/- 1.9,38.2 +/- 10.8,13.5 +/- 0.6,67.4 +/- 4.3
RP,51.1 +/- 2.1,51.0 +/- 1.4,47.3 +/- 1.1,76.4 +/- 5.2,18.1 +/- 0.2,82.6 +/- 0.6
RP CC,42.6 +/- 0.9,42.5 +/- 0.9,41.4 +/- 1.2,42.2 +/- 1.7,17.0 +/- 0.2,81.5 +/- 0.4
\end{filecontents*}
\pgfplotstabletypeset[
results,
images,
every row 2 column 1/.style={highlight},
every row 3 column 2/.style={highlight},
every row 3 column 3/.style={highlight},
every row 2 column 4/.style={highlight},
every row 4 column 5/.style={highlight},
every row 3 column 6/.style={highlight},
every row 4 column 6/.style={highlight},
]{figures/multiple_CIFAR10.csv}
}
\end{table}

\begin{table}[h]
\caption{Metrics on homogenization results in the unsupervised setting after 100 optimization epochs for CIFAR-10. The mean and the standard deviation over 5 runs are given. $n_{\set B}=512$.}
\centering
\scalebox{0.93}{
\footnotesize
\begin{filecontents*}{figures/multiple_CIFAR10_unsupervised.csv}
Methods,acc,acc(val),acc(ver),l2-err,PSNR,SSIM
NN,85.6 +/- 1.3,79.8 +/- 1.7,76.4 +/- 0.3,18.1 +/- 2.3,17.4 +/- 0.7,85.3 +/- 1.0
NN ALL,91.8 +/- 0.6,84.5 +/- 0.7,83.6 +/- 1.0,14.1 +/- 2.3,22.1 +/- 0.6,94.5 +/- 0.3
RANDOM NN,57.1 +/- 6.4,53.2 +/- 4.9,49.3 +/- 5.4,34.3 +/- 17.8,12.1 +/- 0.7,67.2 +/- 3.4
RP,47.7 +/- 1.1,45.8 +/- 1.0,42.8 +/- 0.9,36.1 +/- 5.1,15.5 +/- 0.3,79.0 +/- 0.7
\end{filecontents*}
\pgfplotstabletypeset[
results,
images,
every row 1 column 1/.style={highlight},
every row 1 column 2/.style={highlight},
every row 1 column 3/.style={highlight},
every row 1 column 4/.style={highlight},
every row 1 column 5/.style={highlight},
every row 1 column 6/.style={highlight},
]{figures/multiple_CIFAR10_unsupervised.csv}
}
\label{tab:cifar10results_unsupervised}
\end{table}

\begin{figure}
    \centering
    
    \photostrip{\textit{Ground Truth}}{figures/reconstruction_CIFAR10_ground_truth.png}
    \photostrip{\textit{Distorted}}{figures/reconstruction_CIFAR10_distorted.png}
    \photostrip{CRITERION}{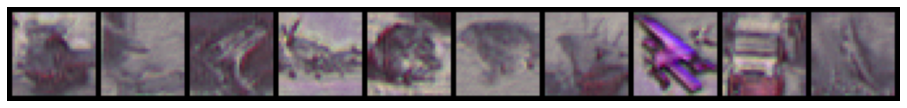}
    \photostrip{NN}{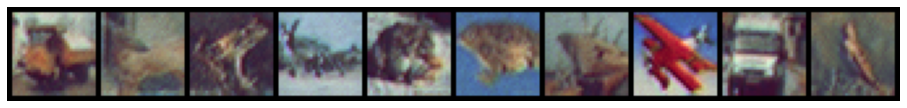}
    \photostrip{NN CC}{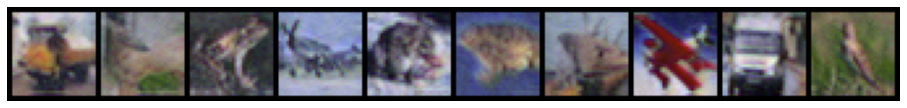}
    \photostrip{NN ALL}{figures/reconstruction_CIFAR10_NN_ALL_epoch_100.png}
    \photostrip{NN ALL CC}{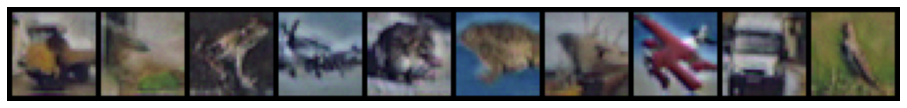}
\end{figure}

\begin{figure}
    \centering
    
    \photostrip{RANDOM NN}{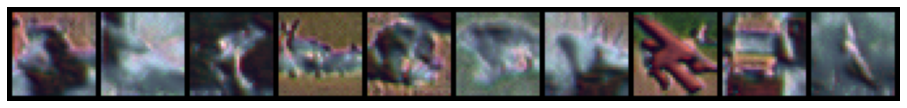}
    \photostrip{RANDOM NN CC}{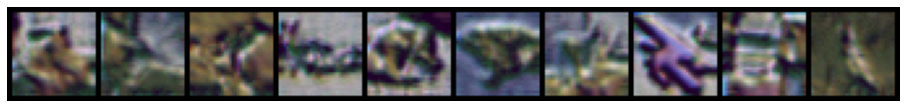}
    \photostrip{RP}{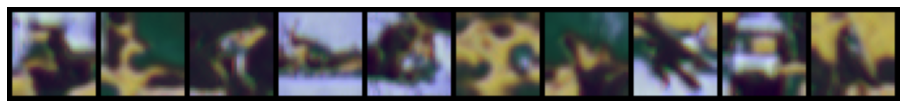}
    \photostrip{RP CC}{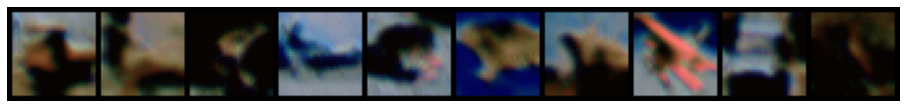}
    
    \caption{Results of the homogenization on CIFAR-10 data set after 100 epochs}
    \label{app:CIFAR_Images}
\end{figure}

\begin{figure}
    \centering
    
    \photostrip{\textit{Ground Truth}}{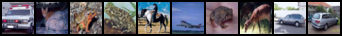}
    \photostrip{\textit{Distorted}}{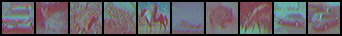}
    \photostrip{NN}{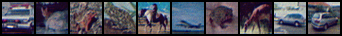}
    \photostrip{NN ALL}{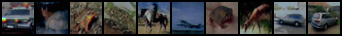}
    \photostrip{RANDOM NN}{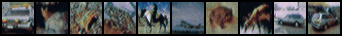}
    \photostrip{RP}{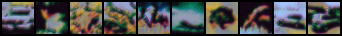}
    
    \caption{Results of the homogenization on CIFAR-10 data set in an unsupervised setting after 100 epochs.}

    \label{fig:CIFAR_Images_unsupervised}
\end{figure}
\newpage
\subsection{MNIST}
\label{app:mnist}

\begin{table}[h]
\centering
\caption{Metrics on homogenization results after 100 optimization epochs for MNIST. The mean and the standard deviation over 5 runs are given. $n_{\set B}=512$.}
\scalebox{0.93}{
\footnotesize
\begin{filecontents*}{figures/multiple_MNIST.csv}
Methods,acc,acc(val),acc(ver),l2-err,PSNR,SSIM
CRITERION,99.8 +/- 0.0,97.7 +/- 0.0,90.4 +/- 0.2,9.8 +/- 0.1,15.7 +/- 0.0,80.6 +/- 0.0
NN,97.5 +/- 0.2,98.0 +/- 0.1,89.7 +/- 1.9,16.9 +/- 0.5,18.0 +/- 0.1,88.5 +/- 0.4
NN CC,97.9 +/- 0.2,98.2 +/- 0.1,97.8 +/- 0.3,14.3 +/- 1.1,20.1 +/- 0.4,91.9 +/- 1.6
NN ALL,98.3 +/- 0.3,98.1 +/- 0.3,98.0 +/- 0.9,9.5 +/- 1.7,21.2 +/- 0.1,90.0 +/- 1.1
NN ALL CC,97.7 +/- 0.3,98.3 +/- 0.2,96.7 +/- 0.4,13.2 +/- 1.1,22.2 +/- 0.1,93.3 +/- 1.0
RANDOM NN,99.8 +/- 0.0,97.7 +/- 0.1,90.3 +/- 0.7,9.9 +/- 0.3,15.8 +/- 0.0,80.7 +/- 0.0
RANDOM NN CC,99.8 +/- 0.0,97.8 +/- 0.1,90.5 +/- 0.4,10.0 +/- 0.2,15.8 +/- 0.0,80.7 +/- 0.0
RP,99.8 +/- 0.0,98.0 +/- 0.1,97.3 +/- 0.1,7.2 +/- 0.1,15.3 +/- 0.0,81.0 +/- 0.0
RP CC,99.8 +/- 0.0,97.9 +/- 0.0,96.6 +/- 0.1,7.6 +/- 0.2,15.4 +/- 0.0,81.0 +/- 0.0
\end{filecontents*}
\pgfplotstabletypeset[
results,
images,
every row 0 column 1/.style={highlight},
every row 5 column 1/.style={highlight},
every row 6 column 1/.style={highlight},
every row 7 column 1/.style={highlight},
every row 8 column 1/.style={highlight},
every row 9 column 1/.style={highlight},
every row 4 column 2/.style={highlight},
every row 3 column 3/.style={highlight},
every row 7 column 4/.style={highlight},
every row 4 column 5/.style={highlight},
every row 4 column 6/.style={highlight},
]{figures/multiple_MNIST.csv}
}
\label{tab:MNIST_results}
\end{table}

\begin{figure}
{
    \centering
    
    \photostrip{\textit{Ground Truth}}{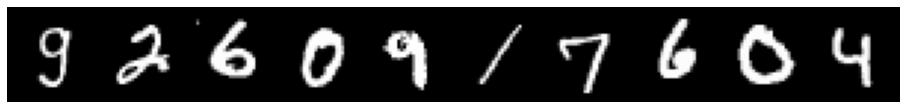}
    \photostrip{\textit{Distorted}}{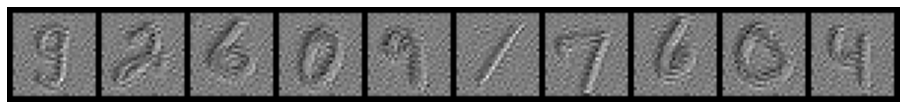}
    \photostrip{CRITERION}{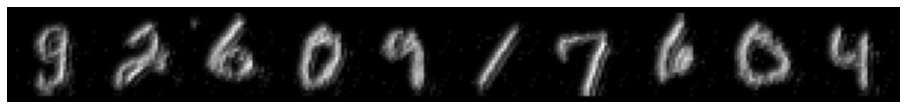}
    \photostrip{NN}{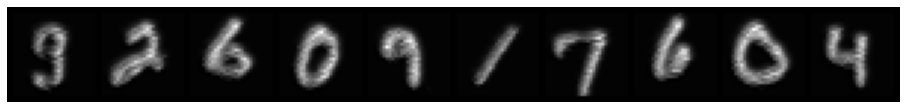}
    \photostrip{NN CC}{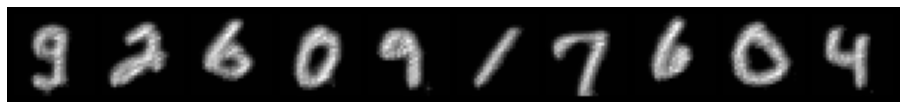}
    \photostrip{NN ALL}{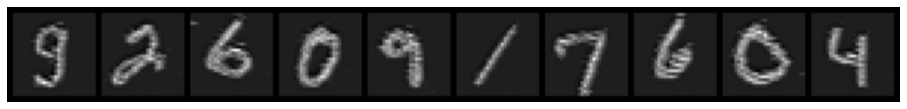}
    \photostrip{NN ALL CC}{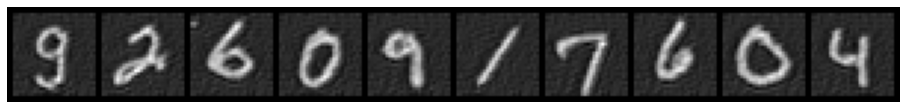}
}
\end{figure}

\begin{figure}
    \centering
    
    \photostrip{RANDOM NN}{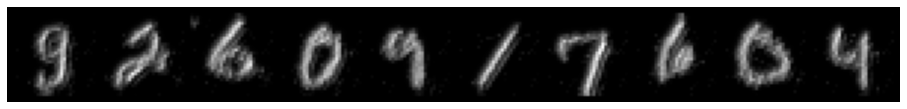}
    \photostrip{RANDOM NN CC}{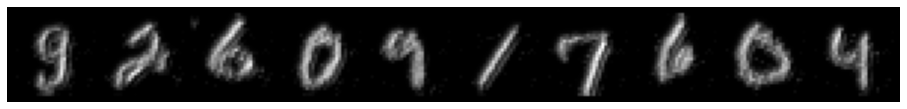}
    \photostrip{RP}{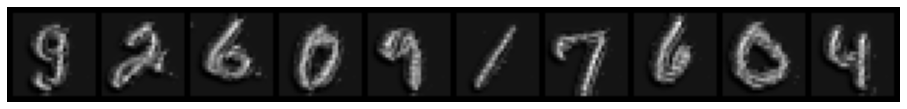}
    \photostrip{RP CC}{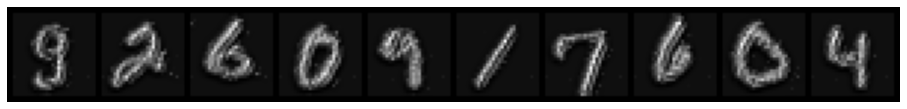}
    \caption{Results of the reconstruction task on MNIST data set after 100 epochs}
    \label{fig:MNIST_Images}
    
\end{figure}

\newpage
\section{Parameter-Settings}
\label{app:hyper}

%
%
\begin{table}[h]
\caption{Parameters and specifications of the experiments. $n_{\set A}$, $n_{\set B}$, $n_{\set C}$ denote the sample size of the source, target and validation data set respectively. $C$ is the number of classes and $d$ the number of input dimensions of the data set. $\kappa$ is the distortion parameter of the perturbation model. $s_\text{width}$ and $s_\text{width}$ are the width and the depth of the residual network used for reconstruction. $s_\text{RP}$ is the number of random projections, the  width used for RP. $r_\text{crit}$ and $r_\text{stats}$ are the loss weighting factors of the objective function.}
\begin{minipage}{0.5\textwidth}
\centering
\caption*{\normalsize{\textbf{MNIST}}}
\pgfplotstabletypeset[
parameters={\textbf{Data Set}},
every row no 3/.style={before row={\\ Properties \\\midrule}},
every row no 5/.style={
    after row={\\ \textbf{Distortion} \\\midrule}
},
every row no 6/.style={
    after row={\\ \textbf{Neural Network} \\ \textbf{Architecture} \\\midrule
        main network: &ResNet20 \\
        verifier network: &ResNet9 \\
    }
},
every row no 7/.style={
    before row={\multicolumn{2}{l}{\makecell[l]{\\ \textbf{Reconstruction Model}}} \\\midrule}
},
every row no 10/.style={
    before row={\\ \textbf{Optimization} \\ \midrule}
},
]{
$n_{\set A}$ &= 60,000
$n_{\set B}$ &= 512
$n_{\set C}$ &= 1024
$C$ &= 10
input shape &= (1, 28, 28)
$d$ &= 784
$\kappa$ &= 0.3
$s_\text{width}$ &= 8
$s_\text{depth}$ &= 8
$s_\text{RP}$ &= 512
learning rate &= 0.1
batch size &= 128
$r_\text{crit}$ &= 1
$r_\text{stats}$ &= 0.001
}
\end{minipage}
%
%
\begin{minipage}{0.5\textwidth}
\centering
\caption*{\normalsize{\textbf{CIFAR10}}}
\pgfplotstabletypeset[
parameters={\textbf{Data Set}},
every row no 3/.style={before row={\\ Properties \\\midrule}},
every row no 5/.style={
    after row={\\ \textbf{Distortion} \\\midrule}
},
every row no 6/.style={
    after row={\\ \textbf{Neural Network} \\ \textbf{Architecture} \\\midrule
        main network: &ResNet34 \\
        verifier network: &ResNet18 \\
    }
},
every row no 7/.style={
    before row={\multicolumn{2}{l}{\makecell[l]{\\ \textbf{Reconstruction Model}}} \\\midrule}
},
every row no 10/.style={
    before row={\\ \textbf{Optimization} \\\midrule}
},
]{
$n_{\set A}$ &= 50,000
$n_{\set B}$ &= 512
$n_{\set C}$ &= 1024
$C$ &= 10
input shape &= (3, 32, 32)
$d$ &= 3072
$\kappa$ &= 0.15
$s_\text{width}$ &= 8
$s_\text{depth}$ &= 8
$s_\text{RP}$ &= 512
learning rate &= 0.1
batch size &= 128
$r_\text{crit}$ &= 1
$r_\text{stats}$ &= 10
}
%
%
\end{minipage}
\end{table}

\end{document}